%% file: main.tex
\documentclass[10pt,journal,twocolumn]{IEEEtran}
\ifCLASSOPTIONcompsoc
  \usepackage[nocompress]{cite}
\else
  \usepackage{cite}
\fi

\ifCLASSINFOpdf
\else
\fi

\usepackage{bbm}
\usepackage{epsfig}
\usepackage{graphicx}
\usepackage{amsmath,amssymb}
\usepackage{eucal,bibspacing}
\usepackage{cite}
\usepackage{cs}
\usepackage{mathsymb}
\usepackage{colortbl}
\usepackage{color}
\usepackage{xcolor}
\usepackage{algorithmic}

\usepackage{multirow}
\usepackage{pbox}
\usepackage{wrapfig}
\usepackage{caption}
\usepackage[linesnumbered,algoruled,boxed,lined]{algorithm2e}

\usepackage{algorithm}

\def\bf{{\boldsymbol f}}

\begin{document}

\title{Learning Resolution-Adaptive Representations for Cross-Resolution Person Re-Identification}

\author{Lin Wu, Lingqiao Liu, Yang Wang, Zheng Zhang, Farid Boussaid, Mohammed Bennamoun

\IEEEcompsocitemizethanks{\IEEEcompsocthanksitem 
\protect\\ L. Wu, F. Boussaid and M. Bennamoun are with The University of Western Australia, Perth 6009, Australia. E-mail: \{lin.wu; Farid.Boussaid; Mohammed.Bennamoun\}@uwa.edu.au. 
\protect\\ L. Liu is with The University of Adelaide, Adelaide 5005, Australia. E-mail: lingqiao.liu@adelaide.edu.au.
\protect\\ Y. Wang is with Hefei Univerity of Technology, Hefei 230009, China. E-mail: yangwang@hfut.edu.cn.
\protect \\ Z. Zhang is with School of Computer Science and Technology, Harbin Institute of Technology, Shenzhen 518055, China. E-mail: darrenzz219@gmail.com.
}
}

\IEEEtitleabstractindextext{%
\begin{abstract}
The cross-resolution person re-identification (CRReID) problem aims to match low-resolution (LR) query identity images against high resolution (HR) gallery images. It is a challenging and practical problem since the query images often suffer from resolution degradation due to the different capturing conditions from real-world cameras. To address this problem, state-of-the-art (SOTA) solutions either learn the resolution-invariant representation or adopt super-resolution (SR) module to recover the missing information from the LR query. This paper explores an alternative SR-free paradigm to directly compare HR and LR images via a dynamic metric, which is adaptive to the resolution of a query image. We realize this idea by learning resolution-adaptive representations for cross-resolution comparison. Specifically, we propose two resolution-adaptive mechanisms. The first one disentangles the resolution-specific information into different sub-vectors in the penultimate layer of the deep neural networks, and thus creates a varying-length representation. To better extract resolution-dependent information, we further propose to learn resolution-adaptive masks for intermediate residual feature blocks. A novel progressive learning strategy is proposed to train those masks properly. These two mechanisms are combined to boost the performance of CRReID. Experimental results show that the proposed method is superior to existing approaches and achieves SOTA performance on multiple CRReID benchmarks.
\end{abstract}

}

\maketitle

\IEEEdisplaynontitleabstractindextext

\IEEEpeerreviewmaketitle

\section{Introduction}\label{sec:intro}

\IEEEPARstart{P}{erson} re-identification (re-ID) is the task of matching the image of the same person across other images taken by different cameras. It is attracting increasing attention due to its wide applications in person tracking \cite{Track-Detection}, surveillance, and forensics \cite{Re-ID-forensics}. A significant volume of existing works in re-ID focus on developing feature representation or metrics that can handle the image variations due to illumination change or occlusion. An important assumption made by all these methods is that both the query and galley images have similar (high) resolutions. However, such an assumption may not hold in real-world scenarios, as the image resolution may vary due to the different distances between cameras and the subject of interest. For example, images captured by surveillance cameras (i.e., the query image) are generally in low resolution (LR), whereas the gallery ones are typically in high resolution (HR). This gives rise to the problem of cross-resolution person re-ID (CRReID). In CRReID, the query image could be a LR image, the challenge is to effectively match it against the HR gallery images.

\begin{figure}[t]
\includegraphics[width=1\linewidth]{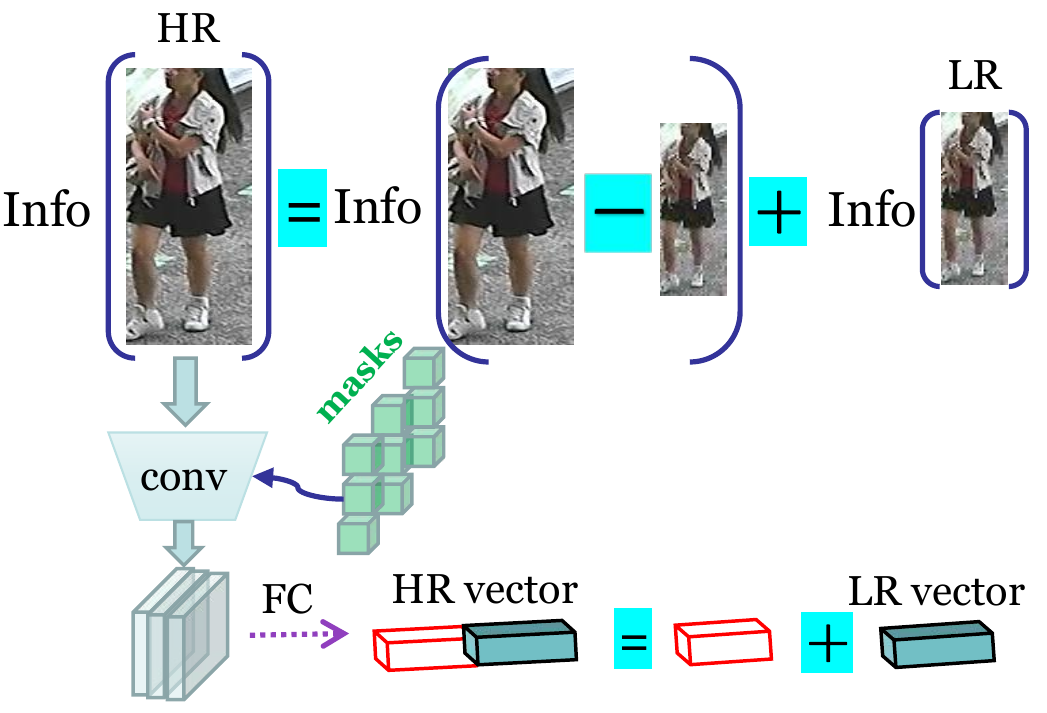}
\caption{The proposed method learns resolution-adaptive representations for CRReID. Through a resolution-adaptive masking and a varying-length representation, we can encode an HR image into the sub-vectors corresponding to the information shared with its LR counterpart, and more sub-vectors corresponding to the extra HR information.}\label{fig:motivation}
\end{figure}


To address the CRReID problem, state-of-the-art (SOTA) methods would employ either methods \cite{SING,CAD-Net,PRI,MRJL,HRNet-ReID} with super-resolution (SR) modules or methods \cite{RAIN} that learn resolution-invariant features The former first recovers the missing details of LR queries before performing the re-ID. The basic assumption is that by using the prior knowledge learned from the training data, the missing details of LR images can be recovered or at least be estimated in a way that will benefit the cross-resolution comparison. However, such a pipeline heavily depends on the recovery output, and yet there is no guarantee that the useful details can be recovered. Moreover, if the input resolution is not seen by the SR model, one cannot properly recover the HR details. The latter tries to learn feature representation that are invariant to resolutions as a way to facilitate cross-resolution comparison. However, such a scheme might also have the risk of losing discriminative information.

This paper explores a different paradigm to directly compare HR and LR images neither relying on super-resolution modules nor seeking invariant features. \textit{The key idea of our approach is to build a metric that is adaptive to the resolution of the query image}. In other words, a query image can select the most appropriate distance metric according to its resolution to compare with the HR gallery images. As shown in Fig. \ref{fig:motivation}, we realize this idea by learning \textit{resolution-adaptive representations}. Specifically, we propose two resolution-adaptive mechanisms (see Section \ref{sec:varying-length representation} and \ref{sec:resolution-adaptive-mask}). The \textbf{first} one is a varying-length image representation with the representation length (dimensionality) determined by the resolution of the image: the higher resolution the input image has, the longer its representation. This mechanism encourages the disentanglement of the features shared across resolution and features specific to HR images.
Specifically, the representation of an HR image is composed of the sub-vectors corresponding to its lower resolution counterpart and extra dimensions corresponding to the higher resolution part. This design captures the essential relationship between HR and LR images -- an HR image should contain all the information of its LR version, and allows images with different resolutions to be comparable via their shared sub-vectors. The \textbf{second} mechanism further strengthens the resolution-adaptive capability by learning resolution-specific masks applied to the intermediate activations of a neural network. Due to the resolution-dependent correlation between the feature blocks, the co-adaptation may occurs if one trains multiple masks in end-to-end \cite{Co-adaptation}. To this end, we design a \textit{progressive} training strategy that can be more effective than the standard end-to-end training strategy to train those resolution-adaptive masks. Through extensive experiments, we show that those two proposed mechanisms are complementary and can be combined to lead to superior performance than the SOTA approaches which are based on super-resolution techniques (see Section \ref{sec:exp}).

Our contributions are summarized below: 1) we propose a varying-length representation to adaptively encode the visual patterns of images from different resolutions, enabling a convenient comparison between images at different resolutions; 2) a resolution-adaptive mechanism is designed by introducing resolution-adaptive masks for intermediate residual feature blocks; 3) a novel progressive training strategy is proposed for training the resolution-adaptive masks.

\input{related}
\input{method}

\input{experiment}

\section{Conclusion}\label{sec:con}
In this paper, we present a novel approach to produce resolution-adaptive representations for cross-resolution person re-identification (CRReID). Specifically, we propose two novel adaptation mechanisms: a varying-length representation learning to produce the feature vector with varied dimensions corresponding to resolution levels, and a set of resolution-adaptive masks applied to intermediate feature blocks to further enhance the resolution disentanglement. The two strategies are combined to achieve the SOTA performance on multiple CRReID benchmarks, especially the merits of addressing the resolution mismatch issue.

\section*{Acknowledgement}
This work was funded by Australian Research Council (Grants DP210101682 and DP210102674). This work was partially funded by NSFC U19A2073, 62002096.

\ifCLASSOPTIONcaptionsoff
  \newpage
\fi



\bibliographystyle{IEEEtran}\small
\bibliography{allbib}

\end{document}

%% file: related.tex
\section{Related Work}

\subsection{Standard Person Re-ID}
Recent person re-ID methods provide person representations that are robust to variations caused by various factors such as human pose, occlusion, and background clutter. For instance, part-based methods \cite{RPP,SpindleNet,PDC,Part-Bilinear,Dual-part-align,MSCAN,Multi-granularity-reID} describe a person image as a combination of body parts either explicitly or implicitly. A number of explicit part-based methods use off-the-shelf pose estimators to extract body parts (e.g., head, torso, legs) with their corresponding features. Instead of explicitly estimating the human pose, implicit part-based methods \cite{RPP,Multi-granularity-reID} rather divide each person image into different horizontal parts with multiple scales. As such, they can exploit the various partial information of the image, and provide a feature representation that is robust to occlusion. Several other approaches \cite{ABD-Net,HA-CNN,HydraPlus-Net,MSCAN} leverage  attention mechanisms to highlight the discriminative parts and remove the background clutter. Other research directions focus on using domain adaptation for person re-ID \cite{SPGAN,Exemplar-reID,GLT,UMDA-reID}. For instance, Zhong \textit{et al.} \cite{Exemplar-reID} proposed to generalize the re-ID model by considering the intra-domain variations of the target domain. Bai \textit{et al.} \cite{UMDA-reID} improved unsupervised domain adaptation (UDA) for re-ID by identifying the domain-specific and domain-fusion views. However, all aforementioned approaches assume that both query and gallery images have similar (high) resolutions, making them not suitable to real-world scenarios.

\subsection{Cross-Resolution Person Re-ID (CRReID)}
To address the practical challenge of CRReID, two main categories of methods have been developed: 1) metric learning or dictionary learning based approaches \cite{SLDL,JUDEA,SDF}; and 2) super-resolution (SR) based approaches \cite{SING,CAD-Net,PRI,MRJL,HRNet-ReID,DDGAN,CRGAN}. For instance, to overcome the resolution mismatch, Jing \textit{et al.} \cite{SLDL} developed a semi-coupled low-rank dictionary learning method to associate the mapping between the HR and LR images. Li \textit{et al.} \cite{JUDEA} introduced a method to jointly perform the cross-scale image alignment and multi-scale distance metric learning. However, all above methods are inherently limited in their matching ability due to the missing details in LR images.

Super-resolution based approaches cope with cross-resolution re-ID via a recovery and re-ID process \cite{Joint-bilateral}. An early work presented by Jiao \textit{et al.} \cite{SING} uses a set of SR sub-networks to improve the compatibility between SR and re-ID. CSR-Net \cite{CSR-GAN} explores cascading multiple SR-GANs \cite{SR-GAN} to progressively recover the details of LR images for resolution alignment. However, these models adopt a separate SR component in the recovery and re-ID pipeline. Instead of applying separate SR models, a novel architecture based on adversarial learning, called RAIN \cite{RAIN}, was proposed to align and extract the resolution-invariant features in an end-to-end fashion. Inspired by RAIN \cite{RAIN}, CAD-Net \cite{CAD-Net} further improves the performance by aligning the distributions between HR and LR images. More specifically, CAD-Net \cite{CAD-Net} jointly considers the resolution-invariant representations and the fine-grained detail recovery in LR input images. However, an outstanding issue remains is that the resolution of the query is unknown to us. To tackle this issue, Han \textit{et al.} \cite{PRI} proposed a framework to adaptively predict the optimum scale factor for the LR images so as to benefit the recovery for the CRReID. Another recent work presented by Cheng \textit{et al.} \cite{MRJL} explores the influence of different resolutions on feature extraction. They show that LR images can provide complementary information to the HR images. Considering the complementary information from the LR images, Cheng \textit{et al.} \cite{MRJL} developed a joint multi-resolution framework based on a reconstruction network and a multi-branch feature fusion network. In contrast, this paper proposes a method that has a few major differences: first, our method \textit{does not need any SR processing} for cross-resolution matching; and second we directly learn resolution-adaptive representations which are amenable for cross-resolution comparison.

%% file: method.tex
\section{Proposed Method}\label{sec:method}

\begin{figure*}[t]
\includegraphics[width=13cm,height=4cm]{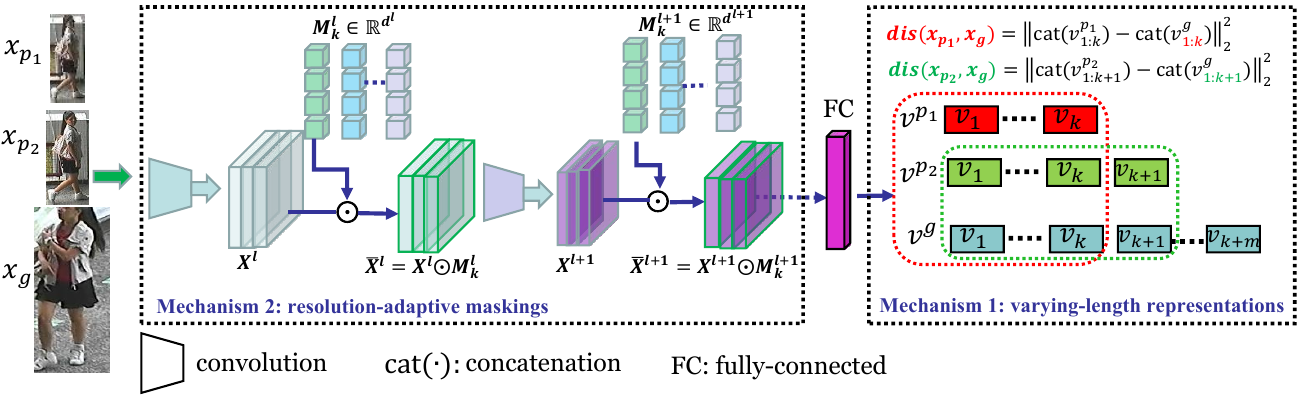}
\caption{The scheme of learning resolution-adaptive representations for person images with different resolutions. Given the query images $x_{p_i} (i=1,2)$, and a gallery image $x_g$ at different resolution levels, we propose two mechanisms to learn resolution-adaptive representations that are convenient for cross-resolution comparison. \textbf{Mechanism 1}: the varying-length representation learning to produce a feature vector with varied dimensions, i.e., the gallery vector $v^g$ contains all sub-vectors that the query vector $v^{p_i}$ has, plus extra dimensions in the HR image. \textbf{Mechanism 2}: resolution-adaptive maskings ($\mathbf{M}^l$) are applied to the $l$-th feature block $\mathbf{X}^l$, yielding $\bar{\mathbf{X}}^l$. The final resolution-adaptive metric is returned by calculating the distance: $dis(x_{p_i},x_g)$. Figure best viewed in color.} \label{fig:two-mechanisms}
\end{figure*}

In this section, we first provide a problem statement and an algorithmic overview of our approach, and then elaborate on components of the proposed method.  Central to our framework are two mechanisms:
\textbf{(1)} a varying-length representation learning to disentangle the resolution-specific information into different sub-vectors; and
\textbf{(2)} a set of learnable resolution-adaptive masks that are applied to intermediate feature blocks at different residual convolutions.

\subsection{Problem Statement and Framework Overview}\label{sec:problem-statement}
We aim to learn a model that can match a low-resolution query image against the high-resolution gallery images. We assume that the resolutions of both the query and gallery images are provided. In practice, the resolution could be estimated from the size (number of pixels) of images or pedestrian bounding boxes since the height of people is relatively fixed. Without loss of generality, we assume that the resolution could be quantized into a set of discrete levels. We denote the resolution with $r$, e.g., $r\in \{1,1/2,1/3,1/4\}$, which is the proportion of the height/width dimension as opposed to the highest resolution considered. For example, if the highest resolution corresponds to $256\times 128$ per person (its resolution ratio $r=1$), for a LR image of size $64\times 32$, its resolution ratio $r =1/4$. In our algorithm, we resize all the images, whether LR or HR, to the size of the highest resolution images via bilinear up-sampling.

Following the setting of CRReID \cite{SING}, we down-sample the HR training images to form various LR images to simulate the LR query images. The aim of our training algorithm is to learn a resolution-adaptive metric, that is:
\begin{align}
    dist(x_p,x_g) = M(x_p,x_g,r_p),
\end{align} where $dist(x_p,x_g)$ returns the distance between a query image $x_p$ and a gallery $x_g$. We implement this similarity measure via a model $M$. An important characteristic of this model is that the resolution ratio of the query image $r_p$ is the input of $M$, and thus the metric is \textit{resolution adaptive}. More specifically, we implement $M$ by \textit{learning resolution-adaptive representations} and we propose two resolution-adaptive mechanisms to realize that. The first is a varying-length representation that uses varying dimensions to encode a query image with different resolutions: the higher resolution, the higher dimension of the representation. Compared with a LR image, a higher resolution one should contain additional dimensions in the representation which depicts the extra information. To further extract resolution-dependent information, we propose the second mechanism: applying resolution-specific masks to the intermediate residual feature blocks. This further enhances the adaptive capability of the network. We illustrate the two mechanisms in Fig. \ref{fig:two-mechanisms}, and the details are introduced in Section \ref{sec:varying-length representation} and \ref{sec:resolution-adaptive-mask}.

\noindent \textbf{Discussions} Comparing with the resolution-adaptive metric, the resolution-invariant metric or representation seems to be a viable solution. However, since the resolution of the query image is not fixed, learning resolution-invariant features will identify discriminative information that are \textit{shared across all resolutions}. Consequently, the information specific to resolutions higher than the lowest one will not be preserved. This inevitably prevent the network from using more information for matching a moderate LR query to HR gallery images.

\subsection{Mechanism 1: Learning Varying-Length Resolution-Adaptive Representations}\label{sec:varying-length representation}

The varying-length resolution-adaptive representation is motivated by the relationship between HR images and LR images: \textit{a HR image should contain all the information conveyed in the LR image, but also extra information from the higher resolution}. Therefore, when comparing a LR image with a HR image, the comparison should adhere to the information shared between them. Note that we do not assume that the unobserved high-resolution information can be recovered from a LR image by leveraging the object prior as the most super-resolution-based CRReID methods. Applying the above idea into representation learning, we propose to disentangle the information that is shared across resolutions and the information that is specific to HR into different dimensions of the feature representation. For example, for a HR image and a LR image, their shared part will be encoded into a sub-vector of the feature representation and the HR-specific part should be encoded into another sub-vector. When one compares a HR image and a LR image, the comparison should only be based on the shared part. In other words, the LR image will in effect be encoded into a representation with lower dimensions (shorter length). In CRReID, a query image could have different resolutions, thus the above strategy will result in different representation lengths, i.e., the higher resolution of the query is, the more information that can be shared with the HR gallery images, and thus the longer dimension of the representation is.

In our implementation, we define $m$ sub-vectors $\{\mathbf{v}_k\}, k=1,\cdots,m$, with $m$ corresponding to the $m$ different levels of resolution. For images with the highest resolution, all $m$ sub-vectors will be concatenated as the final image representation. For the lowest resolution, only the first sub-vector will be used. Formally, the representation of a query image that corresponds to the $k$-th level resolution, (larger $k$, higher resolution), is $\mathbf{z}_p = cat(\mathbf{v}_1,\cdots,\mathbf{v}_k)$ or $\mathbf{z}_p = cat(\mathbf{v}_{1:k})$ in short, where $cat(\cdot)$ denotes concatenation. For HR gallery images, their representations are the concatenations of all $m$ sub-vectors, that is, $\mathbf{z}_g = cat(\mathbf{v}^g_1,\cdots,\mathbf{v}^g_m)$. When a level-$k$-resolution query image is compared against a HR gallery image, the distance is calculated via
\begin{align}\label{eq:vary_length_eq}
    dis(x_p,x_g) = \|\mathbf{z}_p- cat(\mathbf{v}^g_{1:k}) \|_2^2.
\end{align} In other words, the comparison is conducted by only comparing the top-$k$ sub-vectors of $\mathbf{z}_g$ when we know the query image resolution is at level-$k$.


\subsection{Mechanism 2: Resolution-Adaptive Masking}\label{sec:resolution-adaptive-mask}

The above varying-length representation only adaptively constructs the resolution-specific representation in the penultimate layer of the neural network. To extract more resolution-dependent features, we propose a mechanism to inject the resolution information into the earlier layers of a neural network. More specifically, we build our network based on a residual network \cite{Resnet} with learnable masks: one for a resolution level to the activations after each residual block. Each mask is a vector, with each dimension being a real value between 0 and 1. The mask acts as a dimension-wise scaling factor to the feature maps. Formally, let $\mathbf{X}^l \in \mathbb{R}^{d^l \times H^l \times W^l}$ denote the feature maps after the $l$-th residual block. A set of masks $\{\mathbf{M}_k^l \in  \mathbb{R}^{d^l}\}, k = 1,\cdots,m$, are applied to $\mathbf{X}^l$ by $\bar{\mathbf{X}}^l=\mathbf{X}^l \odot \mathbf{M}^l_k$ \footnote{Please note that we DO NOT have any constraints on the structure of those masks, e.g., requiring each dimension of an individual mask corresponding to certain resolutions. For a given layer, we simply allocate one mask for each level of resolution.}, where $\odot$ denotes the element-wise multiplication and $k$ is the resolution-level of the input image. For input images with varied resolutions, different masks will be chosen to determine the final representation. The values of $\mathbf{M}_k^l$ are parameters to be learned. In practice, we reformulate those masks as an channel-wise scaling layer and learn the layer parameters with the network: 
\begin{align}\label{eq:scaling layer}
    \bar{\mathbf{X}}^l=\mathbf{X}^l \odot \big( \sum_k  s^l_k \mathrm{Sigmoid}(\mathbf{M}^l_k) \big),
\end{align}where $s^l_k = 1$ if the input image is at resolution level $k$, otherwise $s^l_k = 0$. $s^l$ could be considered as an input to the network. $\mathrm{Sigmoid}$ is the Sigmoid function converts the real-valued layer parameters $\mathbf{M}^l_k$ into the range between 0 and 1.

\subsection{Varying-length Prediction on Resolution Variations}

The rationale of the varying-length prediction on varied resolution query is to encode the image by reflecting its resolution. Since the LR image is down-sampled from the same HR image, a down-sampled image share with the content with the original HR image but also contains its own characteristic. In this spirit, the feature vector of each image should be characterized by a combination of commonality and resolution-induced characteristics. To this end, we train a classifier consisting of a set of sub-classifiers, such that an image at a resolution looks up the sub-classifiers to adaptively characterise its own features. Specifically, for an image $x$ with its resolution indication $r_k$, we have its deep feature vector $v=[v_1,\ldots, v_m]\in \mathbb{R}^d$ outputted from the last fully-connected layer of the network, where $d$ denotes the dimension and $v_k\in \mathbb{R}^{d_k}$ is a partition of $v$ with the dimensionality $d_k$. Define $\mathbf{w}^i$ as a classifier for one identity (ID), then for all IDs, we have  a set of classifiers $\mathbf{w}=[\mathbf{w}_1,\ldots, \mathbf{w}_m]\in \mathbb{R}^{d\times C}$, where $\mathbf{w}_k\in \mathbb{R}^{d_k \times C}$ is a prototype-based sub-classifier, and $C$ is the number of total IDs. To perform the prediction on the varying-length feature vector $v$, we calculate the ID prediction logits across all IDs via $z^k=\mathbf{w}_k^T v_k\in \mathbb{R}^C$. Since the ID prediction is to classify each image by evaluating the classifier $\mathbf{w}_k$ into the embedding space, the classifier can be interpreted as the prototype closest to the image in the feature space, and assign to an image the ID label of its nearest prototype. Thus, the prototype-based classifier is effective for classifying images based on the closest training prototypes $\mathbf{w}_k$ in the feature space. Finally, the learning objective yielded by the varying-length prediction incrementally updates prototypes to better discriminate the training images with identity labels. The computational algorithm is illustrated in Algorithm \ref{alg:vary-length-representation}.

\begin{algorithm}[t]\small
 \caption{Resolution Adaptive Representation Learning for CRRe-ID.}\label{alg:vary-length-representation}
\textbf{Inputs:} Training images in the form of query, gallery and the resolution level of the query image $\{x_p, x_g, r_p\}$.\\
\textbf{Output:} Model $M$ that extracts varying-length features.
\begin{algorithmic}[1]
\STATE Define $C$ identity classifiers, $\mathbf{W}\in \mathbb{R}^{d\times C}$, one for each identity.
\FOR{$l = 1,2,\cdots, L$}
\STATE Randomly initialize the layer-wise masks $\mathbf{M}^l=\{\mathbf{M}_k^l\}$ at the $l$-th layer, and fix the parameters of $\mathbf{M}^{l'}$,$ \forall l' < l$. (The lower layer index, the closer to the output layer). 
\FOR{$t = 1,2, \cdots T$}
{\STATE 1. Randomly sample a mini-batch of training triplets. Each sample is with the form $\{x_p, x_g, r_p\}$.
\STATE 2. For each triplet, based on the query resolution $r_q$ and the current layer ID $l$ to determine $s_k^l$ for the scaling layer in Eq. \ref{eq:scaling layer}. Perform forward calculation to obtain varying-length representation $\mathbf{z}=cat(\mathbf{v}_{1:r_p})$. 
\STATE 3. Padding zero to $\mathbf{z}$, making its dimension equal to $d$. Apply the class ID loss and verification loss via Eq. \eqref{eq:verif-loss}. Then perform back-propagation.
 }
 \ENDFOR
 \ENDFOR
 \end{algorithmic}
\end{algorithm}

\subsection{ Resolution-Adaptive Representation Training}
Most of the state-of-the-art (SOTA) person re-ID methods train the models with an identity classification loss $\mathcal{L}_{\mbox{cls}}$ (namely ID loss) and a verification loss $\mathcal{L}_{\mbox{verif}}$. We follow this convention to adopt both losses to train the proposed model. In traditional re-ID model training, the standard ID loss is applied to the fixed-length representation. However, our training set comprises HR images and multiple LR counterparts, which are created by down-sampling the HR images with varied resolution levels. This leads to multiple representations when applying the resolution-adaptive mechanisms. To overcome this issue, we propose to apply zero-padding, i.e., concatenating ``0'' to the representation whose dimension is less than the maximal dimension, to convert a varying-length representation to a fixed-length representation. Then a normal identity classification loss can be applied to learn the model. 

The verification loss $\mathcal{L}_{\mbox{verif}}$ is applied to a binary classifier that predicts whether two samples belong to the same class. In our implementation, we start by padding zeros to the varying-length representations of two images, and send their feature vector difference to a multi-layer perceptron (MLP) to make a binary prediction about whether those two samples are from the same class. The loss function is
\begin{equation}\label{eq:verif-loss}
\begin{aligned}
&\mathcal{L}_{\mbox{verif}} = -\sum_{n}^N y_n \log (p(y_n=1\mid \mathbf{v}_{ij})) \\
&+ (1-y_n)\log(1-p(y_n=1\mid\mathbf{v}_{ij})),
\end{aligned}
\end{equation}
where $\mathbf{v}_{ij}=\mathbf{v}_i-\mathbf{v}_j$ denotes the feature difference and $p(y_n=1\mid \mathbf{v}_{ij})$ is implemented with a MLP, e.g., $p(y_n=1 \mid \mathbf{v}_{ij}) = Sigmoid(f(\mathbf{v}_{ij}))$, where $f$ is a MLP mapping $\mathbf{v}_{ij}$ to a scalar. We define $y_n=1$ if two images are from the same class, otherwise $y_n=0$. Our final loss is the weighted summation of both loss terms, that is, $\mathcal{L}_{\mbox{cls}} + \lambda \mathcal{L}_{\mbox{verif}}$, where $\lambda$ is the balance parameter.


\subsubsection{Analysis of the Identity Classification Loss}

In the following part, we present an analysis on the identity classification loss to show why it can benefit the resolution-adaptive metric learning.
Note that the inner product between an identity classifier and a zero-padded representation will only be determined by the inner product of the sub-vectors corresponding to the non-zero parts of the representation. 
Formally, we could consider that 
a classifier is constructed with $m$ parts too, each part corresponding to one sub-vector in the varying-length representation. In other words, $\mathbf{w}=cat(\mathbf{w}_1,\ldots,\mathbf{w}_m)\in \mathbb{R}^{d}$, where $\mathbf{w}_k\in \mathbb{R}^d_k$. So $\mathbf{w}^\top \mathbf{z}_k = cat(\mathbf{w}^c_1,\ldots, \mathbf{w}_k)^\top \mathbf{z}_k$, where $\mathbf{z}_k$ is an image representation with resolution level $k$. ID loss will encourage samples from the same identity class to move closer to the corresponding classifier $\mathbf{w}$ and thus indirectly pulling those features close to each other. Similarly, we could expect our ID loss will pull $\mathbf{z}^k$ and the first $k$-th subvectors of $\mathbf{z}^{k'}~k' > k$ close to each other, which ensuring that images of the same identity but different resolutions become closer under the proposed distance metric Eq. \ref{eq:vary_length_eq}. This can be explained by the following example.


\noindent\textbf{Example:} Suppose we have two images, an HR image and an LR image belonging to the same identity. By using the varying-length representation, the HR image will produce a representation $cat(\mathbf{v}^1_L, \mathbf{v}^1_H)$ while the LR image will produce a representation $\mathbf{v}^2_L$. Suppose the corresponding non-zero part of the classifier of this identity is $cat(\mathbf{w}_L, \mathbf{w}_H)$, the ID loss will make $\mathbf{v}^2_L$ align with $\mathbf{w}_L$ and $cat(\mathbf{v}^1_L, \mathbf{v}^1_H)$ align with $cat(\mathbf{w}_L, \mathbf{w}_H)$. The latter usually implies that $\mathbf{v}^1_L$ should align with $\mathbf{w}_L$. Thus this indirectly encourages that $\mathbf{v}^1_L$ aligns with  $\mathbf{v}^2_L$, through their shared aligning target $\mathbf{w}_L$. This idea is illustrated in Fig.3. 

\begin{figure}[t]
\centering
\includegraphics[width=1\columnwidth]{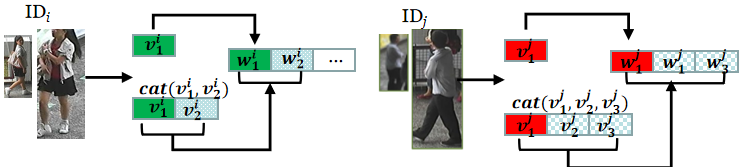}
\caption{By zero-padding, images with the same identity but different resolutions will share the same classifier. This will encourage the shared part (denoted as the red and green blocks) of a HR-image representation and a LR-image representation close to each other if they belong the same identity.}
\end{figure}\label{fig:varying-classifier}




\subsubsection{Progressive Training for Resolution-Adaptive Masks} 

We aim to train the two mechanisms jointly for optimizing the performance. While the most straightforward way is to train them end-to-end via stochastic gradient descent, we empirically find that this standard fashion will lead to compromised performance, probably because of the difficulty of optimization. Alternatively, we propose to introduce the masks at different layers sequentially and train them progressively to avoid the co-adaptation of multiple masks \cite{Co-adaptation}. In our implementation, we start introducing masks to the residual blocks closest to the classifier layer and then gradually add more masks to the residual blocks further away, i.e., from higher to lower. Once new masks are introduced to the residual block, the masks that have been trained with higher-level residual blocks than the current block will be fixed and not updated anymore.  Please refer to the experimental section for more discussion on the effect of this progressive training strategy. The whole training process is shown in Algorithm \ref{alg:vary-length-representation}.

\section{Implementation Details}

The model was implemented using PyTorch. We built the network based on the ResNet-50 architecture with four residual blocks. 
For all resolutions, we resized each image to $256\times 128 \times 3$ and padded with 10 pixels with zero values. Then, we randomly cropped the image into a $256\times 128$ rectangular images. Each image was flipped horizontally with 0.5 probability. The training batch size was 32. The learning rate was set to be 0.00035 and is decayed to $3.5\times 10^{-5}$ and $3.5\times 10^{-6}$ after 40 epochs and 70 epochs, respectively. This warm-up learning rate is shown to be effective to boostrap the network as suggested by \cite{Bag-of-tricks}. We trained the network using ADAM optimizer with 120 epochs in total. The masking mechanism was applied to each residual block, which is a composition of two layers of $3\times3$ conv/batch norm/relu. The last stride of the residual block was set to 1 to achieve a feature map with a higher spatial size ($16\times 8$). For each block, we initialized a resolution mask matrix using Gaussian randomness, followed by a sigmoid function. The sigmoid function restricts the real values in the range of (0,1). The parameter $\lambda$ is empirically set to be 0.5. 

%% file: experiment.tex
\section{Experiments}\label{sec:exp}
In this section, we evaluate the proposed method on several benchmark datasets and compare against SOTA methods. We report both quantitative and qualitative results as well as ablation studies to thoroughly analyze our method.

\subsection{Datasets}
Following existing works \cite{SING,CAD-Net,PRI}, we adopt the multiple low-resolution (MLR) person re-ID evaluation setting on three datasets. The details of each dataset are described as follows. \textbf{(1) CAVIR} \cite{CAVIR} dataset is a real-world dataset composed of 1,220 images of 72 identities and two camera views. Following \cite{SING,PRI,CAD-Net}, we discard 22 identities that only appear in the closer camera. The remaining images of 50 identities are randomly and evenly divided into two halves for training and test. \textbf{(2) CUHK03} \cite{Deepreid} dataset contains over 14,000 images of 1,467 identities captured by 5 pairs of cameras. Following \cite{PRI,CAD-Net}, we adopt the 1,367/100 identities as the training/test split. \textbf{(3) Market-1501} \cite{Market-1501} dataset consists of 32,668 person images of 1,501 identities observed under six different camera views. The dataset is split into 12,936 training images of 751 identities and 19,732 testing images of the remaining 750 identities.


\begin{table*}[t]
\caption{ Comparison with the state-of-the-art models on three datasets ($\%$). Note that ``-" means that the results are not reported. Best results are in boldface. It is clear that our proposed method consistently outperforms existing methods and leads to the overall best performance. }\label{tab:CMC}
\centering
\setlength{\tabcolsep}{1mm}{
\begin{tabular}{|r|ccc|ccc|ccc|} 
\hline
\multirow{2}*{Method}&\multicolumn{3}{c|}{CAVIAR}&\multicolumn{3}{c|}{MLR-CUHK03}&\multicolumn{3}{c|}{MLR-Market-1501}\\
\cline{2-10}
& Rank-1 & Rank-5 & Rank-10& Rank-1 & Rank-5 & Rank-10 & Rank-1 & Rank-5 & Rank-10\\
\hline
JUDEA \cite{JUDEA} &22.0 &60.1 &80.8 &26.2 &58.0 &73.4  &- &- &-\\
SL$D^2$L \cite{SLDL} &18.4 &44.8 &61.2 &- &- &- &- &- &- \\
SDF \cite{SDF} &14.3 &37.5 &62.5 &22.2 &48.0 &64.0 &- &- &- \\
SING \cite{SING} &33.5 &72.7 &89.0 &67.7 &90.7 &94.7 &74.4 &87.8 &91.6\\
CSR-GAN \cite{CSR-GAN} &34.7 &72.5 &87.4 &71.3 &92.1 &97.4 &76.4 &88.5 &91.9\\
RAIN \cite{RAIN} &42.0 &77.3 &89.6 &78.9 &97.3 &98.7 &- &- &-\\
CAD-Net \cite{CAD-Net} &42.8 &76.2 &91.5 &82.1 &97.4 &98.8  &83.7 &92.7 &95.8\\
PRI \cite{PRI} &43.2 &78.5 &91.9 &85.2 &97.5 &98.8 &84.9 &93.5 &96.1\\
INTACT \cite{INTACT} & -& -& -&86.4& 97.4&98.1 &88.1 &95.0& 96.4\\
CRGAN \cite{CRGAN} & 43.1 & 76.5 & 92.3 & 83.4 & 98.1 & 99.1 & 88.1 &95.0& 96.4\\
HRNet-ReID \cite{HRNet-ReID} & 48.2 & \textbf{84.5} & 96.3 & 88.9 & 96.4 & 98.7 & 87.6 & 95.1 &97.0\\
CamStyle \cite{CamStyle} &32.1 &72.3 &85.9 &69.1 &89.6 &93.9 &74.5 &88.6 &93.0\\
FD-GAN \cite{FD-GAN} &33.5 &71.4 &86.5 &73.4 &93.8 &97.9  &79.6 &91.6 &93.5\\
PCB \cite{PCB} &42.1 &74.8 &88.2 &80.6 &96.2 &98.6 &82.6 &92.7 &95.2\\
PyrNet \cite{PyrNet} &43.6 &79.2 &90.4 &83.9 &97.1 &98.5 & 84.1 &93.0 &96.2\\
DDGAN \cite{DDGAN} & 51.2 & 83.6 & 94.4 & 85.7 & 97.1 & 98.6 &- &- &-\\ 
JBIM + OSNet \cite{Joint-bilateral} & 53.1 & 84.0 & 95.2& 88.7 & 97.5 & 99.0 & - & -&-\\
\hline
Ours & \textbf{63.6} & 79.2 & \textbf{96.6} & \textbf{89.2} & \textbf{98.9} & \textbf{99.8} & \textbf{90.1} & \textbf{96.2} & \textbf{97.7}\\
\hline
\end{tabular}
}
\end{table*}

\subsection{Training and Evaluation Protocols}
To train CRReID models, we construct the training set using the original HR images as well as the down-sampled images with the down-sampling rate $r\in \{2, 3, 4\}$. For evaluation, we follow the MLR evaluation protocol suggested in \cite{SING,CAD-Net,PRI}. Specifically, for each HR image from the test set, we randomly choose a down-sampling rate $r\in\{2,3,4\}$ and use the down-sampled images to construct a query set. The gallery images are all HR images with one randomly selected HR image per person. The random data splits are repeated 10 times and the average value is computed for every 10 trials. For the re-ID evaluation, we use the average Cumulative Match Characteristic (CMC) and report the results at rank-1, 5 and 10. In Section \ref{sect:unseen_resolution_test}, we also investigate the generalization of the proposed method to unseen resolutions at the test time. 



\subsection{Comparison Methods} 

We compare the proposed method against three families of SOTA approaches. The first family comprises a number of CRReID methods based on resolution-invariant representation including SING \cite{SING}, RAIN \cite{RAIN}, and CAD-Net \cite{CAD-Net}. The second family is based on the super-resolution module, which is the mainstream approach in CRReID. Methods like CRGAN \cite{CRGAN}, PRI \cite{PRI} and HRNet-ReID \cite{HRNet-ReID} fall into this category. The third family comprises standard re-ID models including CamStyle \cite{CamStyle}, FD-GAN \cite{FD-GAN}, PCB \cite{PCB}, PyrNet \cite{PyrNet} and DDGAN \cite{DDGAN}. For those methods, we directly quote the results from PRI \cite{PRI}, where the authors strictly reproduce the results by using the same training set as our method. For standard re-ID methods, the training set contains the HR images only.

\subsection{Experimental Results}

\subsubsection{Main Quantitative Results}
The comparison results of our method against SOTA re-ID methods on three benchmark datasets are reported in Table \ref{tab:CMC}. For a fair comparison, we do not combine our method with any pre/post processing, e.g., re-ranking \cite{Re-rank-reID} or part-pooling \cite{RPP}, even though these operations can bootstrap the re-ID performance further. From Table \ref{tab:CMC}, we can make the following observations:\textbf{(1)} Comparing with resolution invariant representation learning methods, i.e., SING \cite{SING}, RAIN \cite{RAIN} and CAD-Net \cite{CAD-Net}, our method provides notable improvement across all evaluation metrics. For instance, our method outperforms CAD-Net \cite{CAD-Net} (a leading recovery and re-ID method) by 20.8\%, 7.1\%, and 6.4\% at rank-1 on CAVIR, MLR-CUHK03, and MLR-Market-1501, respectively. This clearly supports our claim on the advantages of resolution-adaptive representations. \textbf{(2)} We observe that our method can also achieve superior performance over super-resolution based approaches, as shown in the comparison against CRGAN \cite{CRGAN} and PRI \cite{PRI}, which are also SOTA methods in CRReID. Table \ref{tab:CMC} shows that our method outperforms those competing methods by a large margin. This demonstrates that adaptive representation learning is a promising paradigm to solve the CRReID problem. \textbf{(3)} Comparing with standard re-ID methods, such as CamStyle \cite{CamStyle} and FD-GAN \cite{FD-GAN}, that can be applied in CRReID, our method still outperforms those competitors. 

\subsection{Ablation Study}
This section performs ablation studies to examine the effectiveness of resolution adaptive representations and the impact of various components of our method.

\begin{table*}[t]
\caption{Investigation on different components of the proposed method. IDE+Verif is the baseline with both identity classification loss and verification loss. Ours (w/o x) means removing the component x from our approach. Ours (w/o val)* denotes that our model with block-wise masks is trained in an end-to-end fashion. Also note that ``val'' denotes the varying-length representation.}\label{tab:component-study}
\centering
\setlength{\tabcolsep}{1mm}{
\begin{tabular}{|l|ccc|ccc|ccc|} 
\hline
\multirow{2}*{Model}&\multicolumn{3}{c|}{CAVIAR}&\multicolumn{3}{c|}{MLR-CUHK03}&\multicolumn{3}{c|}{MLR-Market-1501}\\
\cline{2-10}
\cline{2-10}
& Rank-1 & Rank-5 & Rank-10& Rank-1 & Rank-5 & Rank-10 & Rank-1 & Rank-5 & Rank-10\\
\hline
IDE + Verif & 33.2 & 56.4& 65.2& 76.5 & 95.7& 98.2 & 81.6 & 92.0& 95.1\\
Ours (w/o mask) & 57.4 & 69.5& 75.6& 85.0& 97.2& 98.6 & 85.1 & 94.2&96.5\\
Ours (w/o val) &57.2 & 71.6& 77.2 &85.8 &97.4 &98.7 & 89.3& 95.8& 97.4\\
Ours (w/o val)* &48.7 & 68.7& 75.6&85.0 &97.1 &98.6 & 89.1& 95.6& 97.4\\
Ours (mask+val) & \textbf{63.6} & \textbf{79.2}& \textbf{96.6} & \textbf{89.2} & \textbf{98.9} & \textbf{99.8} & \textbf{90.1} & \textbf{96.2} & \textbf{97.7}\\
\hline
\end{tabular}
}
\end{table*}

\begin{table*}[t]
\caption{Comparison against a naive solution: training $K$ retrieval models, one corresponding to a possible query resolution. Given a query image with resolution level $k$, its corresponding model is picked to perform retrieval. Our results show that our method outperforms this naive solution significantly. }\label{tab:trivial-solution}
\centering
\setlength{\tabcolsep}{1mm}{
\begin{tabular}{|l|ccc|ccc|ccc|} 
\hline
\multirow{2}*{Model}&\multicolumn{3}{c|}{CAVIAR}&\multicolumn{3}{c|}{MLR-CUHK03}&\multicolumn{3}{c|}{MLR-Market-1501}\\
\cline{2-10}
& Rank-1 & Rank-5 & Rank-10& Rank-1 & Rank-5 & Rank-10 & Rank-1 & Rank-5 & Rank-10\\
\hline
Baseline & 18.1 & 44.8 & 56.4 & 83.3 & 96.0 & 97.1 & 85.6 & 94.2 & 96.0\\
Ours (end-to-end) &58.2 & 73.4 & 89.8 & 87.3 & 97.7 & 98.7& 89.5 & 95.8 & 97.4\\
Ours (progressive) & \textbf{63.6} & \textbf{79.2} & \textbf{96.6} & \textbf{89.2} & \textbf{98.9} & \textbf{99.8} & \textbf{90.1} & \textbf{96.2} & \textbf{97.7}\\
\hline
\end{tabular}
}
\end{table*}

\subsubsection{Comparison with A Naive Solution for CRReID}
 One naive solution to realize query-adaptive metric is to build $k$ versions of gallery images, with each one corresponding to a possible level of image resolution. Based on the query image resolution (unseen query resolution could be assigned to one of the nearest resolutions), one can pick the corresponding version of gallery images to make comparison.
As such, it is intriguing to know whether it is necessary to perform resolution-adaptive representation learning. To verify this, we train several baselines on the three datasets. Specifically, for both MLR-CUHK03 and MLR-Market-1501, we first down-sample the HR training images at different resolutions \{1/2,1/3,1/4\} to match the query at the corresponding resolution. Then, we train a baseline with a group of LR query and LR gallery images under the two losses. Since CAVIR is the only real cross-resolution dataset and its query presumably shows $r=1/2$, we down-sample the HR training images to form the LR counterpart. For a fair comparison, we also train our model with the proposed two mechanisms in an end-to-end manner. All the comparison results are reported in Table \ref{tab:trivial-solution}. This could be because our model can use the training samples from multiple-resolutions. We can see that the proposed model (trained in end-to-end or progressive) has an obvious advantage over the baseline on the three datasets. Interestingly, we found that the progressive training method consistently performs better than end-to-end training, with an improvement ranging from 2-5\%.

\subsubsection{Study of Resolution-Adaptive Mechanisms}
One may wonder the relative contribution of the proposed resolution-adaptive mechanisms, i.e., a varying-length feature representation learning and learnable masks for intermediate activations, to the performance of CRReID. To study the impact of each component, we created four variants of our method: \textbf{(1) Ours (w/o mask)}, which removes the learnable mask and only uses the varying-length feature representations. \textbf{(2) Ours (w/o val)}, which does not use the varying-length feature representations but with the learnable masks. \textbf{(3) Ours (w/o val)*}, which also only uses the learnable masks, but trains in an end-to-end fashion rather than adopting the progressive training strategy. \textbf{(4)} A baseline called IDE+Verif is trained using our method but without either learnable masks or varying-length representations. This serves as a baseline for our ablative analysis. The obtained experimental results are shown in Table \ref{tab:component-study}. As seen, each of the components, i.e., the learnable masks, varying-length feature representations and the progressive training plays a critical role, and removing any of them will lead to degraded performance. Their combination leads to the best accuracy. Also, by comparing against the baseline approach, we observe that using either mechanism alone leads to a significant improvement. This illustrates the effectiveness of both mechanisms.

\begin{table}[t]
\caption{Study of loss functions on MLR-CUHK03. Best results are in boldface.}\label{tab:study-loss-function}
\centering
\setlength{\tabcolsep}{1mm}{
\begin{tabular}{|l|ccccc|}
\hline
\multirow{2}*{Method}&\multicolumn{5}{c|}{MLR-CUHK03}\\
\cline{2-6}
&Rank-1 &Rank-5 &Rank-10 &Rank-20 &mAP\\
\hline
Ours  & \textbf{89.2} & \textbf{98.9} & \textbf{99.8} & \textbf{99.9} & \textbf{88.6}\\
Ours w/o $\mathcal{L}_{\mbox{cls}}$ &77.4 &94.9 &97.4 &98.6 &76.8\\
Ours w/o $\mathcal{L}_{\mbox{verif}}$ &76.3 &94.5 &97.0 &98.5 &75.8\\
\hline
\end{tabular}
}
\end{table}

\subsubsection{Study of Loss Functions}
Our network is trained with two types of loss functions, i.e., the identification loss ($\mathcal{L}_{\mbox{cls}}$) and the verification loss ($\mathcal{L_{\mbox{verif}}}$). Thus, it is important to analyze the impact of different loss functions on network training. To this end, we ablate the two loss functions by comparing 2 variants of our model on the MLR-CUHK03 dataset. Table \ref{tab:study-loss-function} reports the ablation study on the loss functions. When the loss $\mathcal{L}_{\mbox{cls}}$ is turned off, our method sees its rank-1 value drop from 89.2\% (with two loss functions) to 77.4\%.  Without the loss of $\mathcal{L_{\mbox{verif}}}$, our model only achieves 76.3\% at rank-1. This demonstrates that both loss functions are crucial in our method. This is consistent with observations in the existing literature \cite{JLML}.


\begin{figure*}[t]
\centering
\includegraphics[width=12cm,height=6cm]{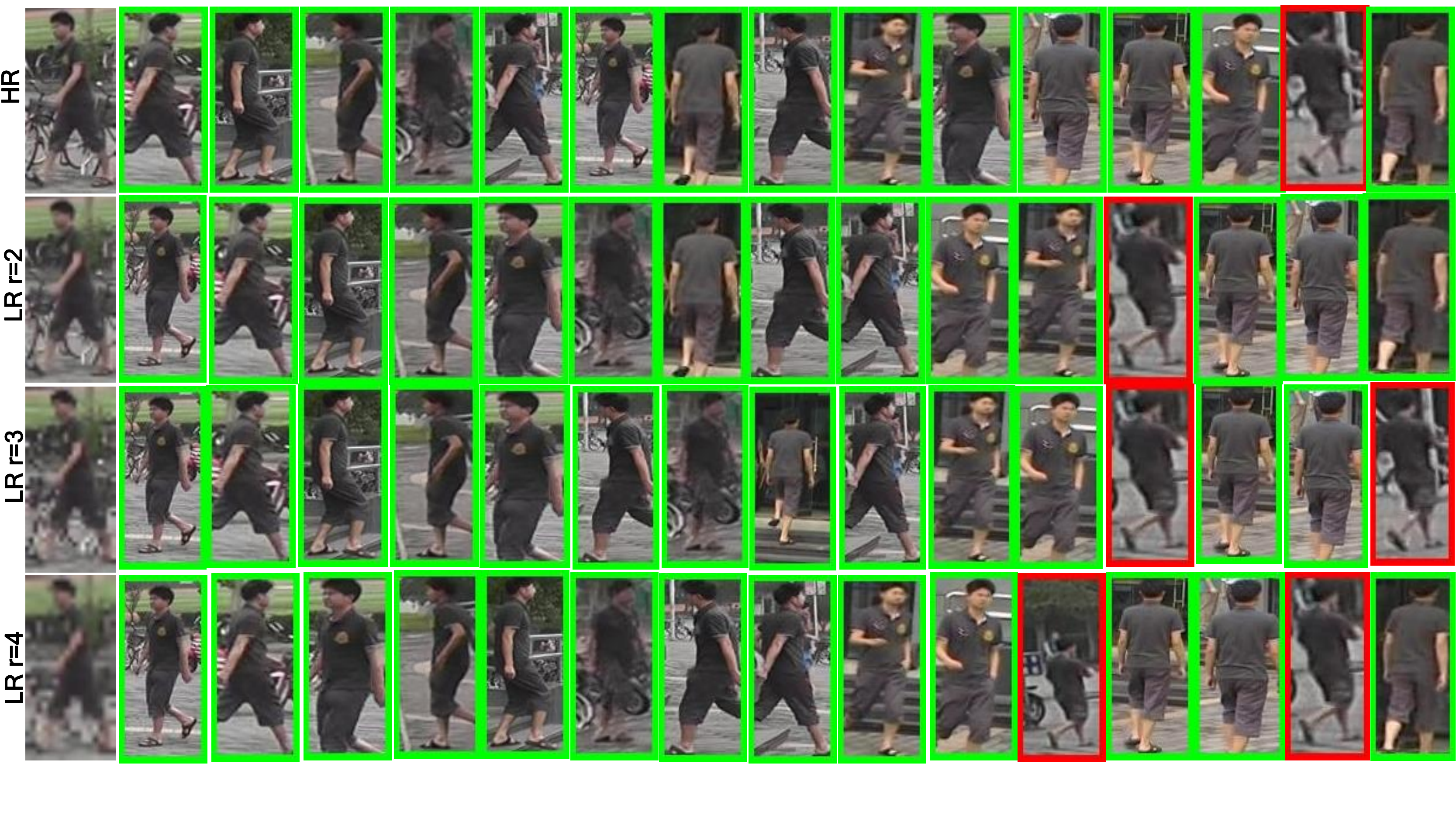} 
\caption{The top-15 ranked gallery images w.r.t the HR query and its down-sampled LR queries with down-sampling rates $r\in \{2,3,4\}$. The ranking evaluation is performed on MLR-Market-1501 dataset. The correct and incorrect matching gallery images are displayed in green and red rectangles, respectively.} \label{fig:top-rank-results}
\end{figure*}


\subsubsection{Top-ranked Gallery w.r.t Varied-resolution Queries} Given a query at different resolutions, we present the first top-15 ranked gallery images from MLR-Market-1501 in Fig. \ref{fig:top-rank-results}. The green and red rectangles indicate the correct and incorrect matches, respectively. The first row of Fig. \ref{fig:top-rank-results} shows the ranking results using the query with its original resolution and matched against HR gallery. When the query has lower resolution, e.g., $r=1/3$ or 1/4, corresponding to the third and fourth rows of Fig. \ref{fig:top-rank-results}, our method can still achieve 13 correct matches out of 15 candidate images for the low-quality queries. This demonstrates the effectiveness of our method in addressing the resolution mismatch between the query and the gallery.

\begin{figure}[hbt]
\includegraphics[width=1\columnwidth]{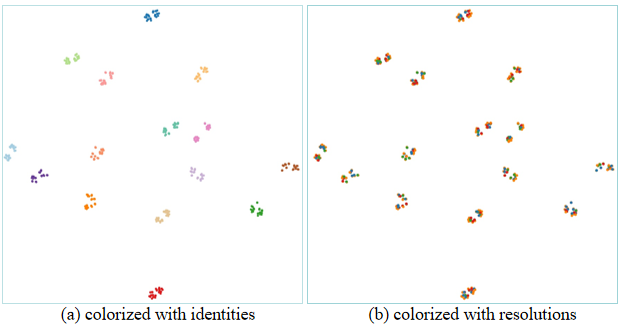}
\caption{The t-SNE visualization of the learned resolution-adaptive features on the MLR-CUHK03 test split. (a) The embedding of identity features. Each color corresponds to one identity. (b) The same data which is colorized to show resolution specifics with four colors corresponding to four down-sampling rates. Best viewed in color.} \label{fig:t-SNE}
\end{figure}

\subsubsection{Embedding of Query-Adaptive Features}
To demonstrate the effectiveness of our method in deriving resolution-adaptive features for different resolutions, we visualize the feature vectors of images from the MLR-CUHK03 \textit{test set} in Fig. \ref{fig:t-SNE}. More specifically, we select 15 different identities, each of which is described by a specific color, and we project the feature vectors in 2D feature space using t-SNE. The projection is shown in Fig. \ref{fig:t-SNE} (a). We observe that our model can establish a well-separated feature space for re-ID. To close up the distribution of different resolutions, we colorize each resolution with a different color in each identity cluster, i.e., four different colors for four resolutions $r\in \{1,2,3,4\}$, and project features vectors via t-SNE. The results are shown in Fig. \ref{fig:t-SNE} (b). Again we observe that the projected feature vectors of the same identity but different down-sampling rates are well separated. The visualization results demonstrate that our method learns resolution-adaptive representations that are effective for cross-resolution person re-ID.

\begin{table*}[t]
\centering
\caption{Study of \textbf{\textit{unseen}} down-sampled resolutions on MLR-CUHK03. The values in brackets indicate the values obtained by turning the unseen resolution into \textcolor{blue}{seen} resolution for training. $x^L_{1/3}\rightarrow x^L_{1/2}$ means assigning the unseen $x^L_{1/3}$ to the down-sampling rate $x^l_{1/4}$ which is seen during training. Note that ``Prog" denotes progressive training and ``E-E" denotes end-to-end. (Best view in color) }\label{tab:test-unseen-resolutions}
\setlength{\tabcolsep}{1mm}{
\begin{tabular}{|cc|cccc|cccc|}
\hline
\multicolumn{2}{|c|}{\multirow{2}*{Test (\textbf{\textit{unseen}})}} &  \multicolumn{4}{c|}{$x^L_{1/3}\rightarrow x^L_{1/2}$ } & \multicolumn{4}{c|}{$x^L_{1/3}\rightarrow x^L_{1/4}$ }\\
\cline{3-10}
&  & Rank-1 & Rank-5 & Rank-10 & Rank-20 & Rank-1 & Rank-5 & Rank-10 & Rank-20\\
\hline
\multirow{2}*{$x^L_{1/3}$} & Prog & 83.4 (\textcolor{blue}{85.6}) & 97.2 (\textcolor{blue}{97.3}) & 98.6 (\textcolor{blue}{98.6}) & 99.3 (\textcolor{blue}{99.6}) & 83.0 (\textcolor{blue}{85.6}) & 96.9 (\textcolor{blue}{97.3}) & 98.6 (\textcolor{blue}{98.6}) & 99.4 (\textcolor{blue}{99.6})\\
 & E-E & 79.5 (\textcolor{blue}{81.5}) & 96.4 (\textcolor{blue}{96.9}) & 98.6 (\textcolor{blue}{98.6}) & 99.3 (\textcolor{blue}{99.3}) & 79.8 (\textcolor{blue}{81.5})& 96.5 (\textcolor{blue}{96.9})& 98.4 (\textcolor{blue}{98.6})& 99.1 (\textcolor{blue}{99.3})\\
\hline
\multicolumn{2}{|c|}{\multirow{2}*{Test (\textbf{\textit{unseen}})}} &  \multicolumn{4}{c|}{$x^L_{1/6}\rightarrow x^L_{1/4}$ } & \multicolumn{4}{c|}{$x^L_{1/6}\rightarrow x^L_{1/8}$ }\\
\cline{3-10}
&  & Rank-1 & Rank-5 & Rank-10 & Rank-20 & Rank-1 & Rank-5 & Rank-10 & Rank-20\\
\hline
\multirow{2}*{$x^L_{1/6}$} & Prog & 80.5 (\textcolor{blue}{84.1}) & 96.4 (\textcolor{blue}{97.2}) & 98.6 (\textcolor{blue}{98.6}) & 99.4 (\textcolor{blue}{99.5}) & 81.0 (\textcolor{blue}{84.1}) & 96.8 (\textcolor{blue}{97.2}) & 98.8 (\textcolor{blue}{98.6}) & 99.6 (\textcolor{blue}{99.5})\\
 & E-E & 76.5 (\textcolor{blue}{79.7}) & 95.7 (\textcolor{blue}{96.6}) & 98.1 (\textcolor{blue}{98.7}) & 98.9 (\textcolor{blue}{99.4}) & 77.9 (\textcolor{blue}{79.7}) & 96.2 (\textcolor{blue}{96.6}) & 98.6 (\textcolor{blue}{98.7}) & 99.2 (\textcolor{blue}{99.4})\\
\hline
\end{tabular}
}
\end{table*}

\subsection{Generalization to Unseen Resolutions}\label{sect:unseen_resolution_test}
In the standard setting for cross-resolution person re-ID, the resolutions or down-sampling rates at the test time are seen during training. In practice, we may encounter the scenario that the test image has a resolution that is not seen during training. It is important that our algorithm could handle this case. To this end, we propose the following scheme: \textbf{(1)} we train our model with a set of fixed down-sampling rates, e.g., $r=\{2, 4, 6, 8\}$. \textbf{(2)} for a test image with unseen down-sampling rate, say $r=3$, we simply assign the test image to the nearest down-sampling rate seen during training and process it as the assigned down-sampling rate. For example, if the resolution of a test image is equivalent to down-sampling the HR image 3 times, we treat the test image as if its $r$ is 2 or 4 and run our algorithm. 

To evaluate the effectiveness of the above scheme, we conduct the following experiment, in which we construct a new MLR dataset on CUHK03 by using the down-sampling rates $r\in \{2,4,8\}$, and then train the model with the progressive and end-to-end mask learning schemes. At the test stage, we consider the query with two unseen resolutions/down-sampling rates, i.e., $r=3$ and $r=6$, denoted as $x^L_{1/3}$ and $x^L_{1/6}$, respectively. Following the principle aforementioned, we can assign it to $r=2 , 4$ or $r=4 , 8$, respectively. We evaluate their performance against baselines that are trained with $r=3$ and $r=6$ images. The results are presented in Table \ref{tab:test-unseen-resolutions}. We can observe that assigning an unseen resolution to the nearest training resolution leads to reasonably good performance. In comparison to the baseline which makes unseen resolution seen during training (i.e., we train a new model with the down-sampling rates $r\in \{2, 3, 6\}$), the performance drop is around 2\% for 1/3 case and is around 4\% for 1/6 case in rank-1. The performance difference becomes much smaller from rank 5 to rank 20. This suggests that this proposed simple solution is sufficient to handle unseen resolutions at test time. Also, we observe that assigning the unseen resolution to a higher resolution neighbour or lower resolution neighbour does not make much difference, the performance is largely comparable in most cases.
